\documentclass[preprint,12pt]{elsarticle}
\usepackage{amssymb}
\usepackage{amsmath}
\usepackage{amsfonts}       % blackboard math symbols
\usepackage[utf8]{inputenc} % allow utf-8 input
\usepackage[T1]{fontenc}    % use 8-bit T1 fonts
\usepackage{graphicx}		% ...
\usepackage{url}            % simple URL typesetting
\usepackage{booktabs}       % professional-quality tables\right 
\usepackage{nicefrac}       % compact symbols for 1/2, etc.
\usepackage{microtype}      % microtypography
\usepackage{array}
\usepackage{xcolor}        % colors
\usepackage{float}			% images and tables [H]
\usepackage{hyperref}       % hyperlinks
\usepackage{multicol}
\usepackage{enumitem}
\usepackage{lscape}
\usepackage{multirow}
\usepackage{hhline}
\usepackage{tabularx}

\usepackage[T1]{fontenc}
\usepackage{lmodern}
\usepackage{microtype}
\journal{Nuclear Physics B}

\begin{document}

\begin{frontmatter}



\title{Profile Generators: A Link between the Narrative and the Binary Matrix Representation}

\author[IDA]{Raoul H. Kutil} %% Author name
%\ead{raoulhugo.kutil@plus.ac.at}
\author[PMU]{Georg Zimmermann}
%\ead{georg.zimmermann@plus.ac.at}
\author[CCNS]{Barbara Strasser-Kirchweger}
%\ead{barbara.strasser-kirchweger@plus.ac.at}
\author[IDA]{Christian Borgelt\corref{cor1}}
\ead{christian.borgelt@plus.ac.at}
\cortext[cor1]{Corresponding Author}

\affiliation[IDA]{organization={Department of AIHI, University of Salzburg},%Department and Organization
            addressline={Jakob Haringer Str. 2}, 
            city={Salzburg},
            postcode={5020}, 
            state={Salzburg},
            country={Austria}}
\affiliation[PMU]{organization={Research Programme Biomedical Data Science, Paracelsus Medical University},
            addressline={Strubergasse 21}, 
            city={Salzburg},
            postcode={5020}, 
            state={Salzburg},
            country={Austria}}
\affiliation[CCNS]{organization={Department of Psychology, University of Salzburg},
            addressline={Hellbrunner Str. 34}, 
            city={Salzburg},
            postcode={5020}, 
            state={Salzburg},
            country={Austria}}

%% Abstract
\begin{abstract}
Mental health disorders, particularly cognitive disorders defined by deficits in cognitive abilities, are described in detail in the DSM-5, which includes definitions and examples of signs and symptoms. A simplified, machine-actionable representation was developed to assess the similarity and separability of these disorders, but it is not suited for the most complex cases. Generating or applying a full binary matrix for similarity calculations is infeasible due to the vast number of possible symptom combinations.\\
The goal of this research is to develop an alternative representation that serves as a link between the narrative form of the DSM-5 and the binary matrix representation and allows for an automated generation of valid symptom combinations.\\
Using a strict pre-defined format of lists, sets, and numbers with slight variations, complex diagnostic pathways involving numerous symptom combinations can be represented. This format, called the symptom profile generator (or simply generator), provides a readable, adaptable, and comprehensive alternative to a binary matrix, while enabling easy generation of symptom combinations. Cognitive disorders, which typically involve multiple diagnostic criteria with several symptoms, can thus be expressed as lists of multiple generators.\\
Representing several psychotic disorders in generator form and generating all symptom combinations showed that matrix representations of complex disorders become too large to manage. The MPCS (maximum pairwise cosine similarity) algorithm cannot handle matrices of this size, prompting the development of a profile reduction method using targeted generator manipulation to find specific MPCS values between disorders.\\
The generators allow for an easier generation of the binary representation and handling of disorders over huge matrices.  Additionally, it makes it possible to calculate a specific case of the MPCS between complex disorders by reducing the number of symptom combinations in an approach called conditional generators.\\
\end{abstract}

%\includegraphics[width=1\textwidth]{figures/VisualAbstract_Gen.png}


%% Keywords
\begin{keyword}
Binary Matrix Representation\sep Symptom Combinations\sep Similarity Measure\sep Cosine Similarity
\end{keyword}

\end{frontmatter}

\section{Introduction}
\begin{figure}[h]
	\centering
	\fbox{\includegraphics[width=0.8\textwidth]{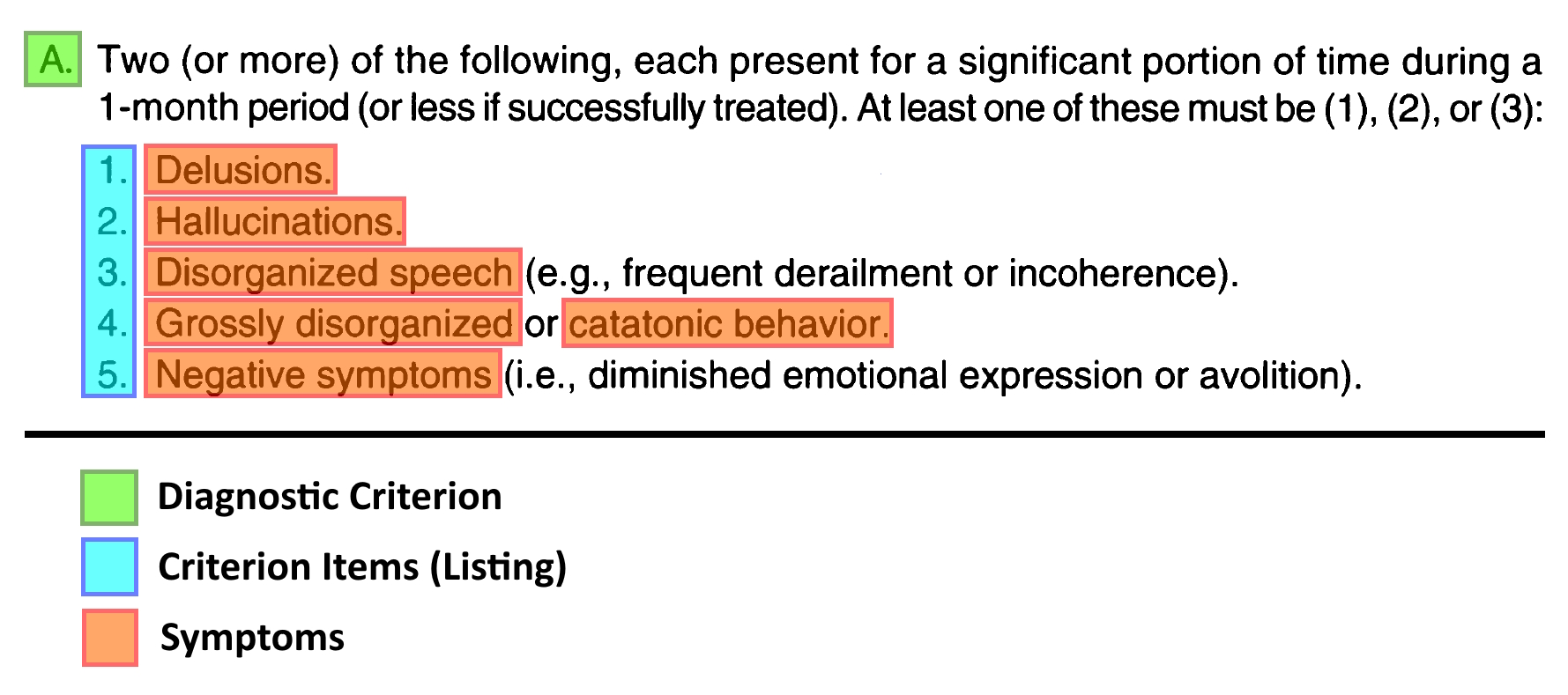}}
	\caption{Diagnostic Criterion "A" of Schizophrenia \citep{dsm5}}
	\label{fig:0}
\end{figure}

Accurate diagnosis is essential for effective treatment, but the process is often complicated by the complexity and symptom overlap among mental disorders. Diagnostic manuals like the Diagnostic and Statistical Manual of Mental Disorders - 5th Edition (DSM-5) and International Classification of Diseases - 10th Edition (ICD-10) provide criteria meant to standardize diagnosis. The DSM-5  is the standard classification system used by mental health professionals to diagnose and categorize mental disorders. It provides clear descriptions, the previously mentioned criteria, and codes to ensure consistency and reliability in diagnosis across clinical settings. The ICD-10 includes much more as it does not specifically focus on mental health issues. Each disorder includes a wide range of possible criteria-satisfying symptom combinations (CSSC) \citep{consenscompute} that fulfil the minimum requirements for diagnosis, though additional symptoms may also be present. %For example, the first criterion for Major Depressive Disorder (MDD) includes nine items, of which any five must be present, with at least one being either depressed mood or loss of interest. 
For example, the first criterion (A) for "Schizophrenia" includes five items, of which any two must be present, with at least one being either "Delusions", "Hallucinations" or "Disorganized speech" (see Figure~\ref{fig:0}).\\

In another case, such as Major Depressive Disorder (MDD), this leads to over seven million possible symptom combinations with the first diagnostic criterion alone. Each combination of symptoms represents a valid diagnostic configuration, as their joint presence is sufficient to establish the corresponding disorder. These will be referred to as symptom profiles, or simply profiles. For Figure~\ref{fig:0} one such profile would be the symptom set \{Delusions, Catatonic Behaviour, Negative Symptoms\}, but recall that this is only one criterion of "Schizophrenia". Such variability makes diagnosis especially difficult when symptoms overlap across disorders, as seen in the shared symptoms of MDD, anxiety disorders, bipolar disorder, and certain personality disorders. This overlap complicates differential diagnosis, the process of distinguishing one disorder from others with similar symptoms. Despite these challenges, accurate diagnosis is crucial, as misdiagnosis can lead to ineffective treatment, prolonged suffering, and increased healthcare costs. While updates to diagnostic criteria, based on clinical research and practitioner feedback, have improved specificity for some conditions, the knowledge remains primarily narrative and not machine-readable. Ontologies like the Human Disease Ontology (HDO) and the ICD ontology aim to structure this information, but they either reflect the authors’ interpretations or fail to represent critical relationships between symptoms and diagnostic criteria. As a result, these tools fall short of supporting scalable computational analysis. The lack of a comprehensive, machine-readable diagnostic framework limits the ability to analyse and distinguish disorders effectively. Clinicians must make decisions under time constraints and cognitive limitations, often relying on heuristics like availability, representativeness, and confirmation bias. While these mental short cuts help manage complexity, they can introduce error, particularly when atypical symptom patterns are involved. To improve diagnostic accuracy and reduce bias, a balance between intuitive judgement and systematic analysis is needed, alongside better tools to represent and navigate diagnostic knowledge.\\

Although prior work has used knowledge graphs to model disease-symptom relationships, ranging from multimodal graph-learning approaches aimed at extracting disease-disease relations (Lin et al., 2023)\citep{Lin2023} to probabilistic knowledge-based models designed for diagnostic inference from symptoms (Jiang et al., 2017)\citep{Jiang2017},our objective differs in that we seek a lightweight, fully text-based representation of symptom combinations rather than a graph-structured or probabilistic framework. Strasser-Kirchweger et al. (2025)\citep{consenscompute} introduces a novel framework using a dichotomous representation (0/1 as indicators) to represent the symptom-criterion relationship. This representation was used to develop a measure to determine the similarity of cognitive disorders as they are described in the DSM-5. First experiments with the new measure aligned with expert insight on the relative similarity between disorders compared to others. At the same time, they showed that complex disorders cannot be directly translated into a dichotomous representation, because the number of symptom combinations is far greater than expected and the algorithm to calculate their similarity cannot handle these disorders in a reasonable amount of time anymore. %These criteria-satisfying symptom combinations (CSSCs) are  referred to as symptom profiles from this point forward. 
In contrast, a recently published dataset on disease-symptom relationships by Ratul et al. (2025)\citep{Ratul2025}, focused on a limited set of diseases and symptoms in a different medical field, employs the same notation but does not require an alternative method for generating symptom combinations, as the total number of possible combinations is under 1,000. By comparison, in the domain of cognitive disorders, major depressive disorder (MDD) is an example where the number of symptom profiles exceeds one billion (1,376,583,579 profiles), making it practically impossible to enumerate all valid profiles without an automated generation method\citep{consenscompute}.\\

This study proposes a new representation that serves as a flexible link between the narrative and binary matrix formats, allowing for easier modifications. Its main objective is to auto-generate the profiles for the binary matrix based on predefined structural and content-based rules that enable a fixed mapping to symptom combinations. Both Bryant et al. (1986)\citep{bdd1986} on Ordinary Binary Decision Diagrams (BDDs) and Minato et al. (1993)\citep{zdd1993} on Zero-Suppressed BDDs show how large combinatorial spaces can be represented compactly using structured decision diagrams, which effectively encode sets of Boolean vectors in a compressed, rule-driven form. However, these techniques rely on diagrammatic structures, whereas we sought a more text-based representation better suited for human-readable specification of constraints and combinations. Our generator-based approach to constructing valid symptom combinations draws on the conceptual foundations of abductive diagnostic reasoning, as formulated by Reggia et al. (1990)\citep{ReggiaPeng1990}. Beyond generating the binary framework, another major challenge is computing the similarity measure, which involves pairwise cosine similarity calculations, which are too numerous to be feasibly computed in regard to the size of these matrices. By adapting this representation, it is, under certain conditions, possible to reduce the number of generated rows while still calculating the same similarity, significantly reducing computational demands. \\
\smallskip

\subsection{The Binary Matrix Representation}
A machine-actionable representation is a translation of narrative text into a format computers can efficiently process. While computers can store and read raw text from sources like the \hbox{DSM-5}, extracting meaningful insights is challenging due to inconsistencies in symptom naming and ambiguities in conjunctions ("and"/"or") from the diagnostic criteria. Natural language processing (NLP) helps interpret these complexities, but a direct translation into a binary system simplifies analysis. A binary matrix is such a machine-actionable representation. In this approach, symptoms are represented as binary values ($1$ for presence, $0$ for absence). The collection of symptoms necessary for a valid diagnosis represented as a list or as a binary vector is called a symptom profile. Two variations are considered: one that focuses solely on symptom presence/absence and another that incorporates the different pathways from the diagnostic criteria for a more detailed representation \citep{consenscompute}.

\subsection{Maximum Profile vs All Profiles}
The "Maximum Profile" method represents each disorder as a single binary vector, marking all symptoms that are present in at least one symptom profile with $1$s and absent ones with $0$s. Vertically stacking these vectors forms a binary matrix with disorders as rows and symptoms as columns. This is called the maximum profile because it maximizes the number of $1$s in the vector for that disorder, capturing all symptoms that can contribute to it in any combination. This approach is highly efficient but fails to capture the nuances and complexity of each disorder; therefore, the following method was developed. The "All Profiles" method accounts for all valid symptom profiles, that is, all symptom combinations fulfilling the diagnostic criteria, forming a separate matrix per disorder. Columns represent symptoms, while rows list the symptom profiles \citep{consenscompute}.\\

The following example illustrates the difference between the AP and MP method and shows the encoding of symptom combinations for both methods into vectors consisting of only $0$s and $1$s. For the two common infections, flu and cold, the base symptom lists differ by only three symptoms.\\
{\small
%[left=0pt, labelsep=0.5em, itemindent=0pt]
\begin{itemize}[left=0pt, labelsep=0.5em, itemindent=0pt, noitemsep, topsep=0pt]
\item \textbf{Flu}:  \{Cough, Runny Nose, Hoarse Headache, Fatigue, Fever, Chills, Nausea\}
\item \textbf{Cold}: \{Cough, Runny Nose, Hoarse Headache, Fatigue\}\\
\end{itemize}}

The union of those sets becomes the base symptom list for the combined binary matrix representation.\\
\renewcommand{\arraystretch}{1.3}
\begin{table}[H]
\advance\tabcolsep-1pt
\begin{center}
\footnotesize
\begin{tabular}{|l|c|c|c|c|c|c|c|c|c|c|c|}
\hline
 & Cough & Runny Nose & Hoarse & Headache & Fatigue & Fever & Chills & Nausea\\
\hline
\textbf{Flu} & 1 & 1 & 1 & 1 & 1 & 1 & 1 & 1\\
\hline
\textbf{Cold} & 1 & 1 & 1 & 1 & 1 & 0 & 0 & 0\\
\hline
\end{tabular}
\vspace{2pt}
\caption{MP representation for Flu and Cold}
\label{table: flu_cold_mm}
\end{center}
\end{table}
In combination with a criterion for both, more diagnostic pathways occur. Thus, the number of symptom profiles (rows) for each sickness increases. 
\begin{itemize}
	\item At least two of the symptoms "\textcolor{blue}{Fever}", "\textcolor{blue}{Chills}" and "\textcolor{blue}{Nausea}"\\ have to be present in a flu.
	\item At least one of the symptoms "\textcolor{red}{Headache}" and "\textcolor{red}{Fatigue}"\\ has to be present in a cold.
\end{itemize}
%\vspace{-20pt}
\renewcommand{\arraystretch}{1.3}
\begin{table}[H]
\advance\tabcolsep-1pt
\begin{center}
\footnotesize
\begin{tabular}{|l|c|c|c|c|c|c|c|c|c|c|c|}
\hline
 & Cough & Runny Nose & Hoarse & \textcolor{red}{Headache} & \textcolor{red}{Fatigue} & \textcolor{blue}{Fever} & \textcolor{blue}{Chills} & \textcolor{blue}{Nausea}\\
\hhline{|=|=|=|=|=|=|=|=|=|}
\textbf{Flu} & 1 & 1 & 1 & 1 & 1 & 1 & 1 & 0\\  \hline
\textbf{Flu} & 1 & 1 & 1 & 1 & 1 & 1 & 0 & 1\\  \hline
\textbf{Flu} & 1 & 1 & 1 & 1 & 1 & 0 & 1 & 1\\  \hline
\textbf{Flu} & 1 & 1 & 1 & 1 & 1 & 1 & 1 & 1\\  
\hhline{|=|=|=|=|=|=|=|=|=|}
\textbf{Cold} & 1 & 1 & 1 & 1 & 0 & 0 & 0 & 0\\ \hline
\textbf{Cold} & 1 & 1 & 1 & 0 & 1 & 0 & 0 & 0\\ \hline
\textbf{Cold} & 1 & 1 & 1 & 1 & 1 & 0 & 0 & 0\\
\hline
\end{tabular}
\vspace{2pt}
\caption{AP representation for Flu and Cold}
\label{table: flu_cold_am}
\end{center}
\end{table}

\subsection{Maximum Pairwise Cosine Similarity (MPCS)}
The MPCS uses the cosine to calculate the similarity of symptom profiles. It determines the highest cosine similarity of a  profile in disorder A for all profiles in disorder B. Each of these values is aggregated into a single value by either the mean or the maximum. The process is repeated with the roles of A and B reversed. Finally, the higher of the two resulting values is selected to represent the similarity between the two disorders \citep{consenscompute}.

\begin{align*}
\mathbf{A},\mathbf{B} \quad &\ldots \quad \textrm{Binary Matrices (Same Number of Columns, i.e. symptoms)}\\
A, B \quad &\ldots \quad \textrm{(Row-)Vectors of Matrix }\mathbf{A},\mathbf{B}
\end{align*}
\begin{align*}
\textrm{Cosine Similarity: }S_C(A,B) &= \frac{\langle A,B\rangle}{\Vert A\Vert_2 \cdot \Vert B\Vert_2}\\
\textrm{Maximum Cosine Similarity: }S_{MC}(A,\mathbf{B}) &= \max\limits_{B \in \mathbf{B}} S_C(A,B)
\end{align*}
\begin{align*}
MPCS(\mathbf{A}, \mathbf{B})= &\max (\phi (\mathbf{A},\mathbf{B}),\phi (\mathbf{B},\mathbf{A}))\\ 
&\textrm{ with} \quad \phi = \phi_{\textrm{mean}} (\mathbf{A},\mathbf{B})= \frac{1}{|\mathbf{A}|} \sum\limits_{A \in \mathbf{A}} S_{MC}(A,\mathbf{B})\\
&\textrm{   or} \quad \phi = \phi_{\textrm{max}} (\mathbf{A},\mathbf{B})= \max\limits_{A \in \mathbf{A}} S_{MC}(A,\mathbf{B})\\ 
\end{align*}

This research uses MPCS$_{\max}$ to specifically refer to the second aggregation method ($\phi_{\max}$), which applies the maximum in the calculation. This approach effectively reduces the entire computation to finding the cosine similarity between the most similar profiles.

\section{Profile Generators}
The one disadvantage of using "All Profiles" (AP) of a disorder in order to make it a machine-actionable representation is that very complex disorders make the number of profiles grow rapidly. While some disorders only have a handful of profiles (e.g. speech sound disorder = 7 profiles ), others can have several hundreds (e.g. schizophrenia = 371 profiles),  but these are still insignificant numbers compared to the most complex disorders that can generate more than a billion profiles (e.g. major depressive disorder = 1,376,583,579 profiles). These large numbers result from the way all profiles from different diagnostic criteria have to be combined. Each criterion contributes a set of valid profiles, which are combined with one profile from each of the other criteria to form the Cartesian product. The binary matrix is formed by stacking these elements vertically.
\begin{align*}
\textrm{Symptom} \quad &\ldots \quad s\\
\textrm{Number of Symptoms, Profiles, Criteria} \quad &\ldots \quad k,l,n \in \mathbb{N}\\
\textrm{Profile} \quad &\ldots \quad p=\{s_1,\ldots,s_k\} \\
\textrm{Criterion} \quad &\ldots \quad C=\{ p_1,\ldots,p_l\} \\
\textrm{All Profiles (AP) / Binary Matrix (BM)}\quad &\ldots \quad \{ C_1 \times C_2 \times \ldots \times C_n\}
\end{align*}

Each criterion may have several symptoms but not all have to be present at the same time. The usual phrasing of such a criteria in the DSM-5 is that at least a certain number of symptoms have to be present, which means every combination of that size %or bigger is a valid profiles that has to be accounted for 
and sets containing these combinations as subsets have to be accounted for (e.g. "one (or more) of the following")\citep{dsm5}. For a set with $n$ symptoms, the number of subsets with $k$ elements is defined by the binomial coefficient, hence the following defines the number of subsets with at least $k$ elements:
\begin{equation}
\sum\limits_{i=k}^{n} \binom{n}{i} = \sum\limits_{i=k}^{n} \frac{n!}{i!\cdot (n-i)!}
\end{equation}

The creation of symptom profiles by hand is an infeasible task due to their complexity (human error) and quantity (millions of profiles) like mentioned above. To solve this problem we developed the so called profile generators, a distinct representation of the diagnostic criteria, that can be used as an input by a mapping algorithm. The generators are essential for translating the formal expressions from the DSM-5 into a format that enables static predefined sets of functions to create the profiles (binary vectors) that form the basis for the binary matrix representation of a disorder. These sets of functions are implemented as enumeration (of the conjunctions of literals of the disjunctive normal form that is the binary matrix) taking generators as input and creating a collection of profiles as an output. This also means that each disorder has its own generator representation. A rule for a specific generator creates a list with sets, whereas each set contains a single valid combination of symptoms. All the sets from different lists get combined with the Cartesian product. In the end, you get a list containing all combinations. The five generators in the following section make it possible to transform all disorders needed for the research in the DSM-5 into the binary matrix representation. All disorder in the DSM-5 follow the same diagnostic criteria guidelines, hence most if not a all of them should be able to transform into a binary matrix with the same five generators.\\

%\subsubsection*{Issues}
%At the same time, symptoms can have slightly different meanings, but count towards the same item in the listing because they stem from a common problem (e.g. fatigue, loss of energy). In this case all such symptoms can also be present in various combinations but only count as one in the criteria rulings. Therefore, each list item can also increase the number of profiles in a significant way. It also happens that symptoms, which are counterparts and should exclude each other, are in the same bullet point (e.g. insomnia, hypersomnia $\Rightarrow$ sleep problems) but for the sake of keeping everything understandable and manageable we make no exceptions in these cases. If such exclusionary symptoms are their own criteria then there is nothing to worry about.\\
%Another issue is that some symptoms have very similar meanings but are expressed differently (e.g., fatigue vs. loss of energy). This topic is addressed in "Machine-actionable criteria chart the symptom space of mental disorders" \citep{consenscompute} and is not relevant for this study. 
%All of the previously mentioned points prevent the creation of the profiles by hand due to their complexity (human error) and quantity (millions of profiles). Having something like an ontology alongside the text can be helpful, but it does not reduce the amount of work required to create profiles. 

\subsection{General Definitions}
A generator is a specifically structured list containing elements like  lists, sets and numbers. The type of elements, their ordering and count specify the individual generator. The generator itself is crucial for mapping the list into a collection of symptom profiles satisfying the diagnostic criteria.

\begin{align*}
\textrm{Symptom} \quad &\ldots \quad s \\
\textrm{Set} \quad &\ldots \quad S \\
\textrm{List} \quad &\ldots \quad L \\
\textrm{Number of Symptoms, Sets} \quad &\ldots \quad n,m \in \mathbb{N}\\
\textrm{Required Number of Symptoms} \quad &\ldots \quad k,r,s,t \in \mathbb{N}_0
\end{align*}

\noindent
\textbf{Generator 0 (G0):}\\
G0 is the identity function for the given symptoms and is used when the symptoms are present in every profile of the critera/disorder. It reproduces the given set exactly, making it a special case of Generator 1 and therefore unnecessary in practice.
\begin{align*}
[S_1]\quad \textrm{with} \quad S_1 = \{s_1, s_2, \ldots, s_n\}
\end{align*}

\noindent
\textbf{Generator 1 (G1):}\\
G1 is used for combinations of symptoms from the set $S_1$ with size greater than or equal to $k$. For $n=k$ follows that G1 = G0. Generates all subsets of the powerset from the input with at least size $k$.
\begin{align*}
[S_1, k]\quad \textrm{with} \quad S_1 = \{s_1, s_2, \ldots, s_n\} \textrm{ and } k<n
\end{align*}

\noindent
\textbf{Generator 2 (G2):}\\
G2 is used for combinations of symptoms from all sets with size greater than or equal to $k$, but counts elements from the same set only once. Generates all subsets of the power set from the combined input sets with at least size $k$, but only counts symptoms from different initial sets towards $k$.
\begin{align*}
[S_1, \ldots, S_m, k]\quad \textrm{with} \quad S_i = \{s_{i,1},\ldots, s_{i,n_i}\} \textrm{ and } k \leq m
\end{align*}

\noindent
\textbf{Generator 3 (G3):}\\
G3 creates disjoint combinations by pairing each individual set in $L_1$ with each individual set in $L_2$. 
It makes the Cartesian product of both lists, which generates all combinations of sets from both lists where no two sets in the combination come from the same list. 
\begin{align*}
[L_1, L_2]\quad \textrm{with} \quad L_1 &= \{S_{1,1},\ldots, S_{1,n_1}\}\\
L_2 &= \{S_{2,1},\ldots, S_{2,n_2}\}
\end{align*}

Using an empty set $\emptyset$ in $L_2$ allows to generate the sets from $L_1$ separately from each other. (e.g. [\{$S_1$, $S_2$, $S_3$\}, \{$\emptyset$\}] $\rightarrow$ \{$S_1$\}, \{$S_2$\}, \{$S_3$\})\\
%$[L_1, L_2] = \{ S_1 \cup S_2 | \exists S_1' \in L_1: S_1 \subseteq S_1' \land \exists S_2' \in L_2: S_2 \subseteq S_2' \}$

The necessity of G3 is highly uncommon and was not observed in any of the disorders tested. Nonetheless, at least one disorder from the DSM-5 (Obstructive Sleep Apnea Hypopnea)\citep{dsm5} requires its application. Even then, there may exist less optimal approaches to avoid employing G3, as its inclusion greatly increases the complexity of both the formalization process and the algorithm presented in Section 2.5 (Conditional Generators).\\

\noindent
\textbf{Generator 4 (G4):}\\
G4 is a necessary extension of G2 because some disorders have sub conditions in their criteria. The tuple $(r,s,t)$ functions like the $k$ in G2, with $r$ and $s$ being the minimum requirement of symptoms for each list, while $t$ is the minimum requirement for the entire generator, which is both lists combined.
\begin{align*}
[L_1, L_2, (r,s,t)]\quad \textrm{with} \quad 
&L_1 = \{S_{1,1},\ldots, S_{1,n_1}\}\\
&L_2 = \{S_{2,1},\ldots, S_{2,n_2}\}\\
\textrm{and} \quad &r\leq n_1, \quad s \leq n_2, \quad t\leq n_1+n_2
\end{align*}

\subsection{Mathematical Definitions}
The following introduces basic set theory concepts needed to understand the formal mathematical definitions of the generators.\\
{\small
\begin{align*}
\textrm{Powerset: } \mathrm{ps}(S) &= 2^S = \{R \subseteq S\} \\
\textrm{Size Filter: } \mathrm{sf}(S,m) &= \{R \in S \mid |R| \geq m\} \\
\textrm{Powerset Extended: } \mathrm{ps^*}(S) &= \{\mathrm{sf}(\mathrm{ps}(R),1) \mid R \in S\} \\
\textrm{Union Product: } \mathrm{up}(S) &= \Big\{\bigcup_{R_i \in S} r_i \ \big|\ \forall i : r_i \in R_i\Big\} \\
\textrm{Union Product Extended: }  \mathrm{up^*}(S) &= \bigcup_{R \in S} \mathrm{up}(R)
\end{align*}}
%\textrm{Union Flatten: } \mathrm{uf}(S) &= \bigcup_{R \in S} R \\

The powerset function generates the full powerset of a given symptom set. To ensure relevance, a size filter is then applied, which removes all subsets that do not meet the minimum threshold $\mathrm{m}$. Building on this, the extended powerset function constructs the powersets of all sets (excluding the empty set) and applies the size filter to each of them. The union product is defined as the Cartesian product of the elements within the specified sets, while the extended union product goes one step further by merging the Cartesian products that originate from the same overarching set. The following examples with arbitrary symptoms ($a$ to $f$) illustrate how these functions work.\\
{\small
\begin{flushleft}
\begin{minipage}{\textwidth}
$\mathrm{ps}(\{a,b,c\}) = \emptyset,\{a\},\{b\},\{c\},\{a,b\},\{a,c\},\{b,c\},\{a,b,c\}\\ \\
\mathrm{sf}(\{ \{a\},\{a,b\},\{a,b,c\},\{a,b,c,d\}  \},3)= \{a,b,c\},\{a,b,c,d\}\\ \\
\mathrm{ps^*}(\{\{a,b\},\{c,d\}\})= \{a\},\{b\},\{a,b\},\{c\},\{d\},\{c,d\}\\ \\
\mathrm{up}(\{\{\{a,b\},\{c\}\}, \{\{d,e\},\{f\}\}\})= \{a,b,d,e\},\{a,b,f\},\{c,d,e\},\{c,f\}\\ \\
\mathrm{up}(\{\{\{\{a,b\},\{c\}\}, \{\{d\}\}\}, \{\{\{a,b\},\{c\}\}, \{\{e\}\}\})= 
\{a,b,d\},\{c,d\},\{a,b,e\},\{c,e\}$
\end{minipage}
\end{flushleft}}
\vspace{12pt}\noindent
While the underlying functions are relatively straightforward, the specification of input, particularly their format, can be a source of confusion. Notably, the powerset function operates on a basic symptom set, whereas the remaining functions require nested sets (lists) as input. Accordingly, the generators are defined in the following way:

\begin{itemize}[left=0pt]
	\item Generator 1 ($G_1$):\\ $\quad [S,k] = \mathrm{sf}(\mathrm{ps}(S), k) \quad$ [ includes Generator 0 ($G_0$) for $k = |S|$ ]

    \item Generator 2 ($G_2$):\\ $[S_1, \ldots, S_n, k] = \mathrm{up^*}(\mathrm{sf}(\mathrm{ps}(\mathrm{ps^*}(\{S_1, \ldots, S_n\})), k))$ 

    \item Generator 3 ($G_3$):\\ $[\{S_{1,1}, \ldots, S_{1,n_1}\}, \{S_{2,1}, \ldots, S_{2,n_2}\}] = \mathrm{up}(\{\{S_{1,1}, \ldots, S_{1,n_1}\}, \{S_{2,1}, \ldots, S_{2,n_2}\}\})$

    \item Generator 4 ($G_4$):
    \[
    [\{S_{1,1}, \ldots, S_{1,n_1}\}, \{S_{2,1}, \ldots, S_{2,n_2}\}, (r,s,t)] = 
    \]
    \[
    \mathrm{up^*}\big( \mathrm{sf}(\mathrm{up}(\{\mathrm{sf}(\mathrm{ps}(\mathrm{ps^*}(\{S_{1,1}, \ldots, S_{1,n_1}\})), r),\phantom{\}), t )\big)}
    \]
    \[
    \phantom{\mathrm{up^*}\big( \mathrm{sf}(\mathrm{up}(\{} \mathrm{sf}(\mathrm{ps}(\mathrm{ps^*}(\{S_{2,1}, \ldots, S_{2,n_2}\})), s) \}), t )\big)
    \]
\end{itemize}
\smallskip
\phantom{....}
\subsection{Implementation of the Generator Mappings (Python)}
\textbf{Generator 0} simply copies the given set without generating additional ones. \textbf{Generator 3} utilizes the predefined Cartesian product function (itertools.product) to create exclusive sets.\\ \textbf{Generators 1, 2,} and \textbf{4} follow a similar approach and are responsible for generating the most symptom combinations. The following is a brief summary of how the implementation uses the generators, with reference to the detailed description in the previous section:
\begin{itemize}
	\item \textbf{G1} generates a power set with elements of a specific size ("combinations()" and "chain.from\_iterable()" in Python)
	\item \textbf{G2} builds on \textbf{G1} by first generating the power set  and then filtering symptom combinations based on the specific requirements.
	\item \textbf{G4} extends \textbf{G2} by applying the power set to all sets in both lists before filtering according to set and list constraints.
\end{itemize}
This procedure generates a lot more symptom profiles than necessary but should be more efficient than generating individual profiles that follow the constraints, because the creation and filtering processes are two separate parts.\\

To better understand how these generators work, the following table illustrates how several sample generators with placeholder symptoms (a, b, c, d, e, f) are transformed into a list of sets.
\renewcommand{\arraystretch}{2}
\begin{table}[H]
\advance\tabcolsep-1pt
\centering
\footnotesize
\begin{tabular}{|c|l|p{6cm}|}
%\begin{tabularx}{\textwidth}{|c|l|X|}
\hline
\textbf{Gen} & \textbf{Example} & \textbf{Result}\\
\hline
G0 & [ \{a,b,c,d\} ] &\{a,b,c,d\}\\
G1 & [ \{a,b,c\}, 2 ] &\{a,b\}, \{a,c\}, \{b,c\}, \{a,b,c\}\\
G2 & [ \{a,b\}, \{c,d\}, \{e,f\}, 2 ] &\{a,c\}, \{a,d\}, \{a,e\}, \{a,f\}, \{b,c\}, $\ldots$ \{a,b,c,d,f\}, \{b,c,d,e,f\}, \{a,b,c,d,e,f\}*\\
G3 & [ [\{a,b\}, \{c\}], [\{d,e\}, \{f\}] ] &\{a,b,d,e\}, \{a,b,f\}, \{c,d,e\}, \{c,f\}\\
G4 & [ [\{a,b\}, \{c\}], [\{d\}, \{e,f\}], (1,0,3) ] &\{a,c,d\}, \{a,c,e\}, \{a,c,f\}, \{b,c,d\}, $\ldots$ \{a,b,c,e,f\}, \{a,b,c,d,f\}, \{a,b,c,d,e,f\}*\\
\hline
\end{tabular}
\vspace{2pt}
\caption{Mappings from specific examples (*full result in appendix)}
\label{table: generator_examples}
%\end{center}
\end{table}
%G3 (with $\emptyset$) & [ [\{a,b\}, \{c\}, \{d\}], [\{\}] ] &\{a,b\}, \{c\}, \{d\}\\
%\vspace{-15pt}
The two generators G2 and G4 in the examples of Table \ref{table: generator_examples} create the most sets (even for simple  configurations). G2 generated 54 sets and G4 generated 33 sets in total.
\smallskip

\subsubsection{Criteria to Generator}
It is essential to see the translation of formal expressions from the \mbox{DSM-5} into the generator form. The following two examples show that small differences in the formulation of the diagnostic criteria can change which generator has to be used for mapping to profiles.
\begin{figure}[H]
	\centering
	\fbox{\includegraphics[width=0.9\textwidth]{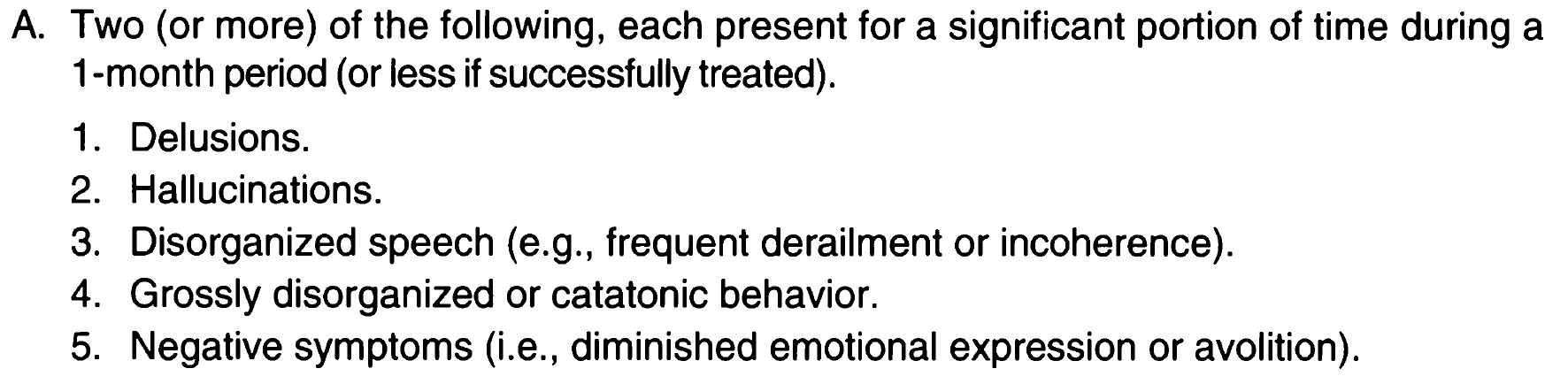}}
	\caption{Schizophrenia - Diagnostic Criteria A (Fake) - Example for G2}
	\label{fig:1}
\end{figure} \vspace{-20pt}
{\footnotesize
\begin{align*}
\textrm{Schizophrenia\_CritA}=[\textrm{ }&\textrm{\{Delusions\}},\textrm{\{Hallucinations\}},\textrm{\{Disorganized Speech\}},\\&\textrm{\{Grossly Disorganized Behaviour, Catatonic Behaviour\}},\\&\textrm{\{Negative Symptoms\}}, 2\textrm{ }]
\end{align*}}

The criterion is transformed into the generator "Schizophrenia\_CritA" (Generator 2 template) by grouping each symptom from the criteria items into individual sets. These sets are then combined into a list, followed by a number indicating the required count of symptoms from the individual sets.\\

\begin{figure}[H]
	\centering
	\fbox{\includegraphics[width=0.9\textwidth]{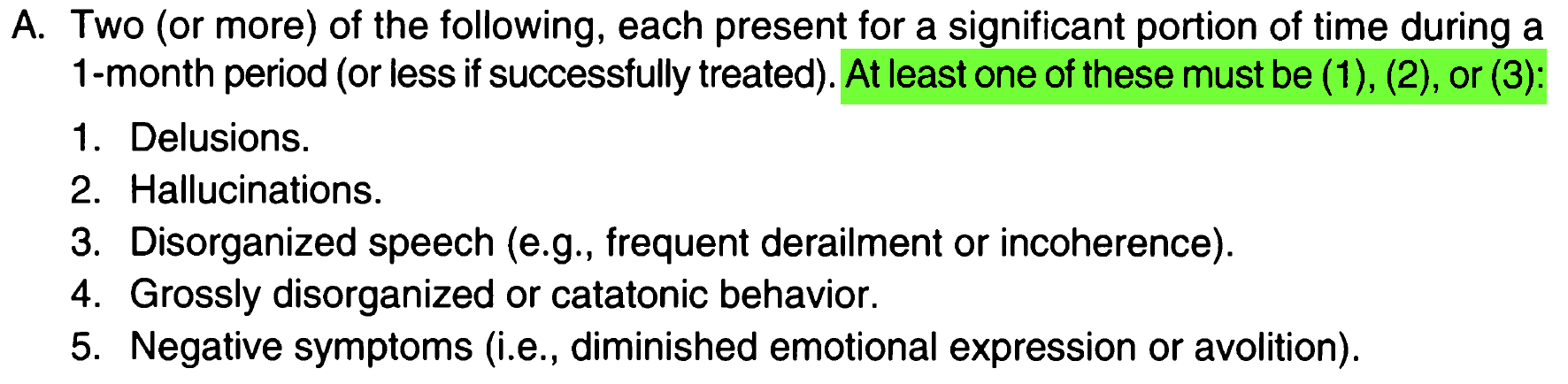}}
	\caption{Schizophrenia - Diagnostic Criteria A (Real) - Example for G4}
	\label{fig:2}
\end{figure} \vspace{-20pt}
{\footnotesize
\begin{align*}
\textrm{Schizophrenia\_CritA}=[\textrm{ }[
&\textrm{\{Delusions\}},\textrm{\{Hallucinations\}},\textrm{\{Disorganized Speech\}}],\\ 
[&\textrm{\{Grossly Disorganized, Catatonic Behaviour\}},\\ &\textrm{\{Negative Symptoms\}}], (1,0,2)\textrm{ }]
\end{align*}}

The criterion is transformed into the generator "Schizophrenia\_CritA" (Generator 4 template) by grouping each symptom from the criteria items into individual sets. The sets are grouped and organized into two individual lists based on the sub-criterion (marked in green). Both lists are put into the overarching generator list but the number at the end is replaced by a 3‑tuple, where the first two elements specify the required number of symptoms from each list (and individual sets), and the third element indicates the total number of symptoms needed.\\

\subsection{Major Depressive Disorder}
The major depressive disorders (MDD) is one of the most complex disorders based on its definition by the diagnostic criteria. The diagnostic criteria span over five parts from point A to E. While point A to C represent a major depressive episode, point D and E try to connect the depressive episode to the major depressive disorder by excluding other disorders that could also cause and explain it. The chapter "Depressive Disorders" of the DSM-5 consists of 8 disorder definitions in total (two of them are just "other specified" or "unspecified" depressive disorders) and they all share symptoms to varying degrees. Especially psychotic disorders have an overlap with depressive disorders as they can cause depressive episodes and share several symptoms in the diagnostic criteria. The diagnostic criteria of psychotic disorders include items with synthetic symptoms that rule out depressive disorders, and vice versa (e.g. symptoms with "$\ldots$ not better explained by $\ldots$"). This illustrates a clear correlation between these two different disorder groups. The definition of the diagnostic criteria for MDD is as follows:
\begin{figure}[H]
	\centering
	\fbox{\includegraphics[width=1\textwidth]{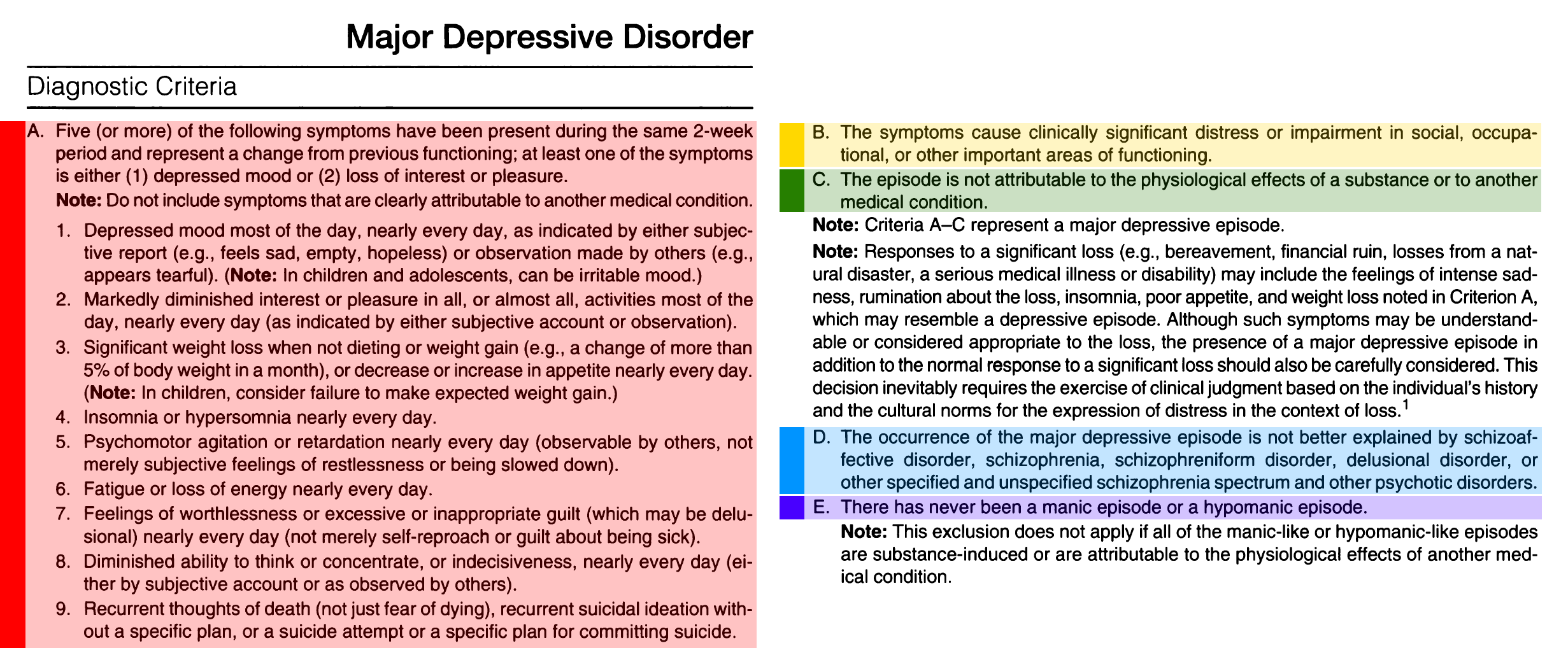}}
	\caption{Major Depressive Disorder - Diagnostic Criteria}
	\label{fig:3}
\end{figure}
All the diagnostic criteria (A - E) have to be translated into generators while also taking the "Notes" into consideration. The first challenge is to identify which generator fits the diagnostic criteria.
\vspace{5pt}
{\small
\begin{align*}
\textrm{Criterion A} \quad &\rightarrow \quad \textrm{Generator 4}\\
\textrm{Criterion B} \quad &\rightarrow \quad \textrm{Generator 1}\\
\textrm{Criterion C} \quad &\rightarrow \quad \textrm{Generator 0}\\
\textrm{Criterion D} \quad &\rightarrow \quad \textrm{Generator 0}\\
\textrm{Criterion E} \quad &\rightarrow \quad \textrm{Generator 1}
\end{align*}}

Then the symptoms are extracted from the DSM-5 (uniformly and consistent) and transformed into sets. The sets are put into the chosen generator template and the quantity identifiers have to be set if necessary (G1, G2 and G4). The individual generators are joined in a list via the union product (modification of the Cartesian product) which becomes the definitive representation of the MDD in the generator form.

\begin{figure}[H]
	\centering
	\fbox{\includegraphics[width=1\textwidth]{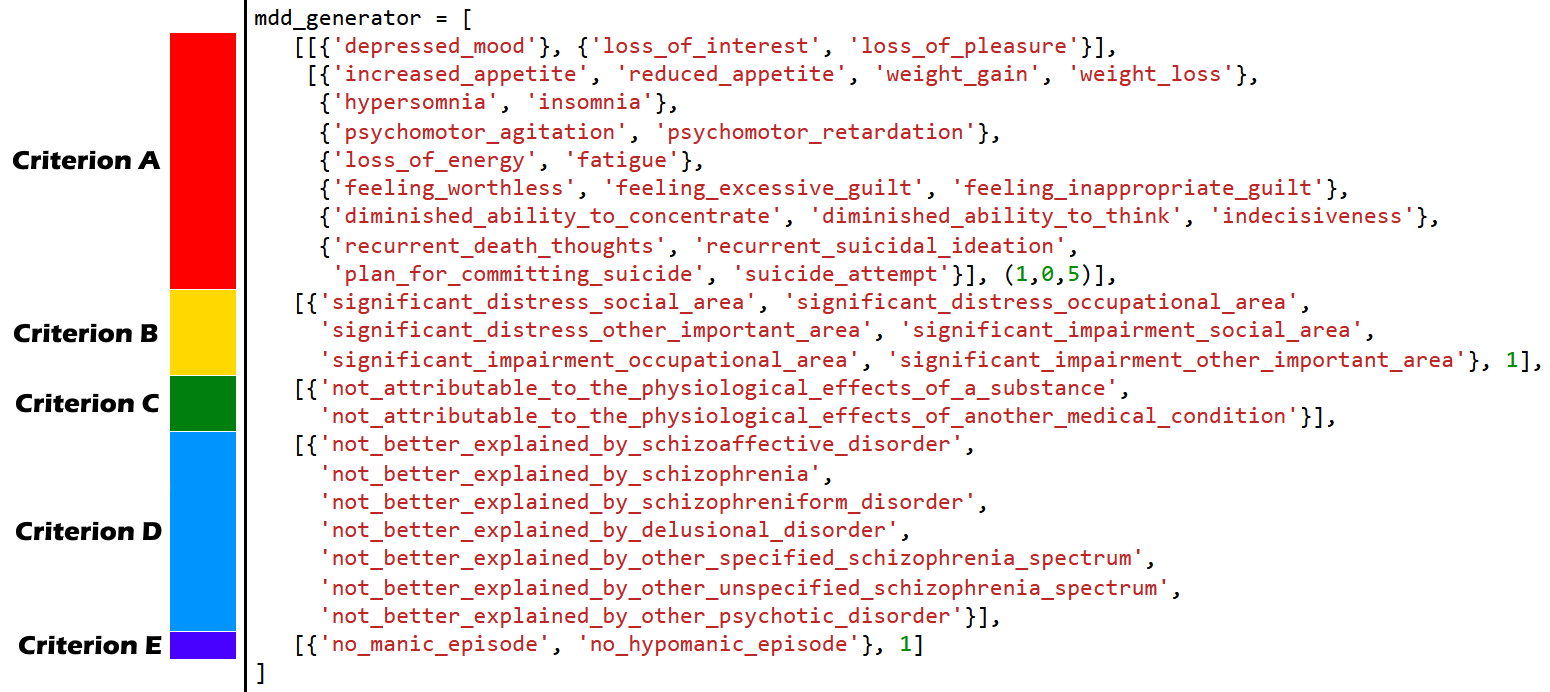}}
	\caption{Major Depressive Disorder - Generator Form}
	\label{fig:4}
\end{figure}
The major depressive disorder will generate $1,376,583,579$ profiles in this fashion. This disorder is certainly an outlier in terms of complexity but not an isolated case in the DSM-5. Another example of such a large number is the panic disorder which generates $3,119,485,608$ profiles. In order to compare disorders of such magnitude with an algorithm that uses any kind of pairwise computation there is a need for a simplification process beforehand.
\medskip
\\

\subsection{Conditional Generators - A Simplification for the MPCS Algorithm}
The generator mapping (to a binary matrix) enables a direct input of two generators to calculate their MPCS, but the size of those matrices (complex disorders = many profiles) statistically increases the runtime of the algorithm. A major portion of the runtime is used for the matrix-matrix multiplication, which can generate a matrix that is to big to handle for common computers. This issue is already addressed in the MPCS source code by switching to matrix-vector multiplication for each row. Although this increases runtime, it avoids the need to hold an excessively large matrix in memory for further computations. However, the problem persists when the matrix-vector multiplication itself becomes too large. There are several ways to optimize the base algorithm.\\
%\begin{itemize}
% 	\setlength\itemsep{0.0em}
%	\item Mathematical Optimization
%	\item Software/Code Optimization (C Implementation, Popcount, etc.)
%	\item Row Reduction or Truncation
%	\item Parallel Computing (Parallelization)
%	\item Using GPUs (CUDA)
%\end{itemize}

%The mathematical and software optimization are done in parallel by using the most basic functions for only the necessary computations. Benchmark tests were done to determine the fastest methods for exemplary data. Even though the python implementation already uses functions that resort to C subroutines, a pure C implementation using the popcount method in conjunction with bit operations could decrease the runtime tremendously.\\ The popcount method is a built-in function (on modern computers) using only CPU instructions to count the number of 1-bits. The denominator is just the square root of the product of both vector's popcount. The inner product in the numerator is computed by applying the logical AND operation on the vectors, and then taking the popcount of the result.\\ Parallelization is also feasible, as matrix multiplication is a prime example of parallel computing. However, it is better to first reduce the number of rows in each matrix before pursuing that approach.\\
To improve computational efficiency, several strategies were considered, including mathematical optimization, software/code optimization, row reduction or truncation, parallel computing, and the use of GPUs (CUDA). Among these, only the mathematical and selected software optimizations were fully implemented, focusing on minimizing redundant computations and improving runtime performance. Benchmark tests were conducted to systematically evaluate and refine these implementations. The two approaches, parallelization and GPU-based computing, were primarily explored at a conceptual level and are currently undergoing experimental evaluation. They are being implemented in the low-level language C using binary vectors (with digits represented by bits) and logical operations. In the prototype implementation developed in the high-level language Python, both approaches would likewise offer substantial performance improvements. However, a row-reduction method remains beneficial regardless of whether these techniques are applied. For this reason, the row-reduction approach was favoured and further developed in the subsequent phase.\\

A problem of the MPCS algorithm is the need for pairwise vector similarity calculation on the fully available matrix representations but this is only necessary for MPCS that uses the mean aggregate function of the highest cosine similarities. If the maximum aggregate is required, the computation can be significantly simplified by constructing a dependent generator pair, two original generators modified to depend on each other with the maximum aggregate in mind. It can be thought of as a cross-dependence lowest common denominator for generators. Consequently, the generators influencing the matrix size the most should be considerably simplified, thereby decreasing their impact on the overall matrix size. The simplification procedure reduces the size particularly for highly similar symptom configurations and in cases where disorders share only a few symptoms.\\

There is no guarantee that the base set of a generator overlaps with just one generator of the other disorder. This turns each generator into a complex case analysis, which can be resolved through sufficient case differentiation. The type of generator also influences how easily the case analysis can be solved. Generators G1, G2 and G4 are responsible for the final profile count, but they can be simplified easily depending on the opposing disorder. In contrast, generator G3 is not a major contributor to the size, but makes the case analysis very complex.
This is the reason why in the simplification process only symptoms from sets of generators that do not appear in G3 (from both disorders) are used.
Both symptom lists can be segmented in three parts:
\begin{enumerate}
\setlength\itemsep{0.0em}
	\item Maximize the number of symptoms (required in both disorders)
	\item Minimize the number of symptoms (required only in the other disorder)
	\item No simplification (G3 in either disorder)
\end{enumerate}
\begin{figure}[H]
	\centering
	\includegraphics[width=0.7\textwidth]{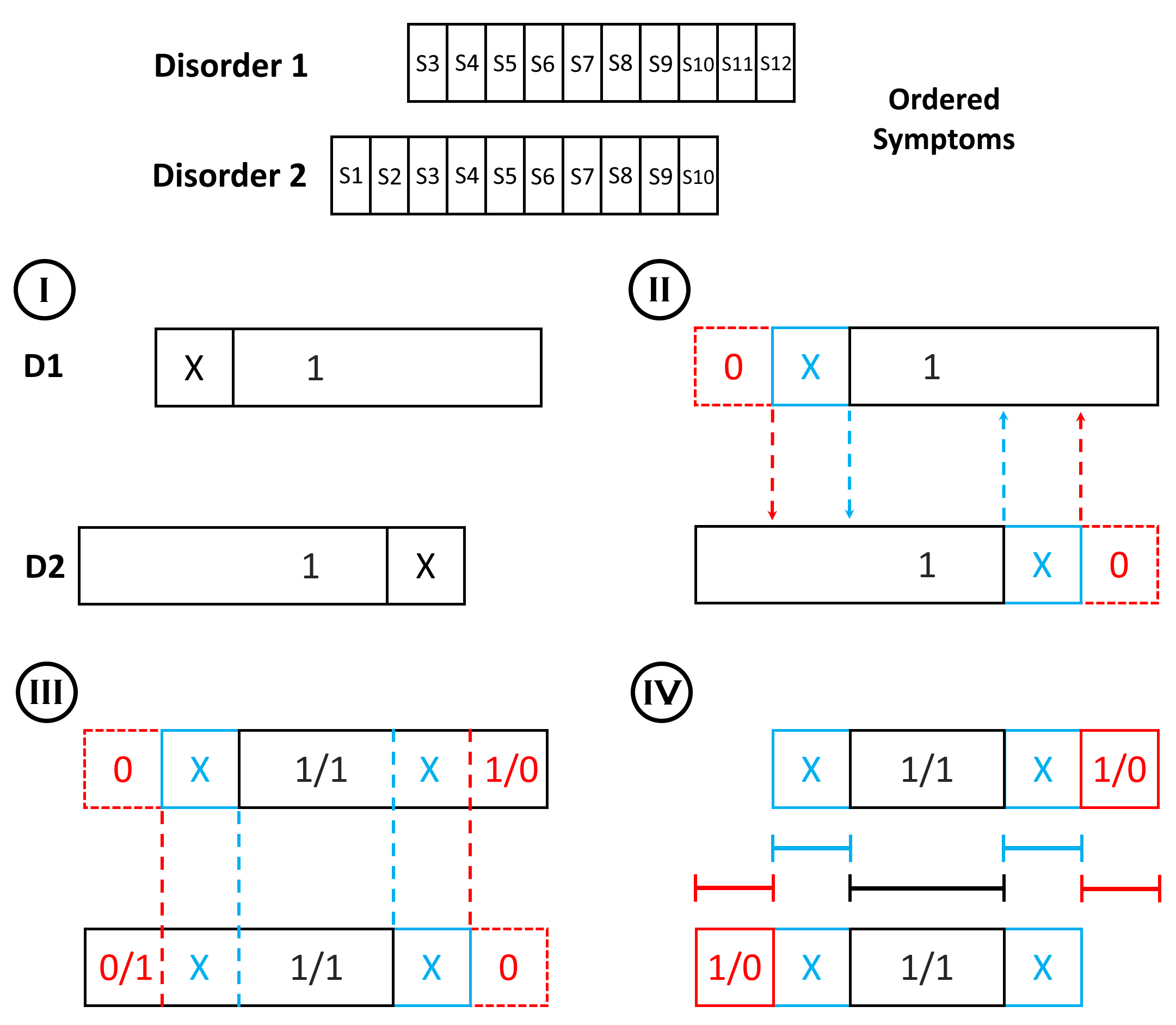}
	\caption{Segmentation process to get three distinct symptom ranges}
	\label{fig:5}
\end{figure}

In Figure \ref{fig:5}, disorder 1 requires symptoms 3 trough 12, while disorder 2 requires symptoms 1 through 10 (in MP representations). The segmentation process divides the symptoms into three distinct groups, each requiring a different approach (maximize [1/1], minimize [1/0], neither max or min [X]) because of the generator they are associated with in either disorder.
This is just a simplified visual of the segmentation process with no initial overlaps of the different groups.\\

\subsection{Conditional Generator Definitions}
The following definitions introduce the basic notations required to understand the construction of the conditional generators.
\begin{align*}
p_i \quad &\ldots \quad \textrm{Symptomsets (Profiles)}\\
M_A=\{p_1,\ldots,p_{n_A}\} \quad &\ldots \quad \textrm{Set of Profiles for Disorder A}\\
M_B=\{p_1,\ldots,p_{n_B}\} \quad &\ldots \quad \textrm{Set of Profiles for Disorder B}
\end{align*}
\begin{align*}
X,Y \quad &\ldots \quad \textrm{Symptomsets}\\
X \cap Y \quad &:= \quad \{x \cap y \mid x \in X, y \in Y\}\\
X \backslash p \quad &:= \quad \{x\backslash p \mid  x \in X\}
\end{align*}
%X \cap p \quad &:= \quad X \cap \{p\}\\

The relevant symptom sets\citep{consenscompute}, denoted by $m^{A+},m^{B+}$, as well as the necessary symptom sets (i.e., symptoms that appear in every profile), denoted by $m^{A-},m^{B-}$, are defined as follows:
\begin{align}
m^{A+}=\bigcup_{p \in M_A} p &,\quad m^{B+}=\bigcup_{p \in M_B} p\\
m^{A-}=\bigcap_{p \in M_A} p &,\quad m^{B-}=\bigcap_{p \in M_B} p
\end{align}

Next, we define the symptom set $p^{*}$, which represents the set of symptoms relevant to both disorders. This set serves as the basis for constructing the disorder-specific symptom sets $p^{*}_A$ and $p^{*}_B$, which additionally include the necessary symptoms unique to each disorder. Subsequently, corresponding sets containing all profiles but excluding these symptoms are also constructed.
\begin{align}
p^{*}= &m^{A+} \cap m^{B+}\\
p^{*}_A= p^{*} \cup m^{A-} &,\quad p^{*}_B= p^{*} \cup m^{B-}\\
M^{-}_A = M_A \backslash p^{*}_A &, \quad M^{-}_B = M_B \backslash p^{*}_B
\end{align}
%M^{\cup}=\bigcup_{p \in M_A \cap M_B} p

The profiles that maximize $p^{*}_A$ and $p^{*}_B$ are collected into corresponding sets, each specific to the respective disorder.
\begin{align}
M^{max}_A = \{p \in M_A: |p \cap p^{*}_A|=\max_{a \in M_A}|a \cap p^{*}_A| \}\\
M^{max}_B = \{p \in M_B: |p \cap p^{*}_B|=\max_{b \in M_B}|b \cap p^{*}_B| \}
\end{align}

We then determine the minimal cardinality of the subsets that are relevant but non-essential using the following expression:
\begin{align}
a_{min} = \min_{a \in M_A, c \in M^{-}_A}|a \cap c|\\
b_{min} = \min_{b \in M_B, c \in M^{-}_B}|b \cap c|
\end{align}

Finally, we construct the set of all profiles that simultaneously maximize and minimize specific subsets of the symptom lists.
\begin{align}
M^{max/min}_A = \{p \in M^{max}_A:\exists a \in M^{-}_A: |p \cap a| = a_{min}\}\\
M^{max/min}_B = \{p \in M^{max}_B:\exists b \in M^{-}_B: |p \cap b| = b_{min}\}
\end{align}

Thus, the profiles left in each set are precisely those that define the similarity value of MPCS$_{\max}$, because they constitute the most similar symptom combinations between the disorders.
\smallskip

\subsection{Finding the Most Similar Profiles}
Our aim is to identify, for two disorders $A$ and $B$, the pair of profiles that are most similar with respect to the cosine similarity measure\citep{salton1983imr}:
\begin{align*}
\text{S}_C(p_A, p_B) = \frac{|p_A \cap p_B|}{\sqrt{|p_A| \cdot |p_B|}},
\end{align*}
where $p_A \in \text{profiles}(A)$ and $p_B \in \text{profiles}(B)$. Consequently, we seek to find
\begin{align*}
(p_A^*, p_B^*) = \arg\max_{\substack{p_A \in \text{profiles}(A) \\ p_B \in \text{profiles}(B)}} \text{S}_C(p_A, p_B).
\end{align*}

%The similarity measure is clearly maximized if $|C_A \cap C_B|$ is as large as possible and $|C_A - C_B|$ as well as $|C_B - C_A|$ are as small as possible; that is, if the presence of symptoms referred to by the diagnostic criteria of both disorders, that is, of symptoms in $S^* = \text{dom}(\text{csscs}(A)) \cap \text{dom}(\text{csscs}(B))$, is maximized and the presence of symptoms exclusive to diagnostic criteria of either disorder is minimized. Since $\text{csscs}(A)$ and $\text{csscs}(B)$ are both upper sets (as shown above), any $C_A \in \text{csscs}(A)$ with $S_{AB} \not\subseteq C_A$ can be extended to a $C'_A \in \text{csscs}(A)$ with $S^* \subseteq C'_A$ and analogously for $B$. Hence for the similarity maximizing pair $(C_A^*, C_B^*)$ of symptom combinations it must hold $C_A^* \cap C_B^* = S^*$.\\
The similarity measure is maximized when the overlap $|p_A \cap p_B|$ is as large as possible, while the non-shared parts $|p_A - p_B|$ and $|p_B - p_A|$ are minimized. In other words, similarity increases when symptoms included in the diagnostic criteria of both disorders, those in $S^* = \text{domain}(\text{profiles}(A)) \cap \text{domain}(\text{profiles}(B))$, are fully present, and symptoms unique to either disorder are minimized. %Because $\text{profiles}(A)$ and $\text{profiles}(B)$ are upper sets (as established earlier),
Any $p_A \in \text{profiles}(A)$ that does not yet contain $S^*$ can be expanded to some $p_A' \in \text{profiles}(A)$ such that $S^* \subseteq p_A'$, and the same holds for $B$. Therefore, for the pair  $(p_A^*, p_B^*)$ that maximizes similarity, it must hold that $p_A^* \cap p_B^* = S^*$.\\
\smallskip

\subsection{Illustrative Example}
The following example presents two disorders, $A$ and $B$, and demonstrates the operation of the conditional generators. While $A$ has the symptom base $\{a,b,c,d,e\}$, $B$ has the symptom base $\{d,e,f,g,h\}$. Both share the same criterion of needing atleast three symptoms for a valid profile.\\

\textbf{State 0}:
\begin{align*}
\textrm{A = [\{a, b, c, d, e\}, 3]} \quad &\ldots\quad 16 \textrm{ profiles}\\
\textrm{B = [\{d, e, f, g, h\}, 3]} \quad &\ldots\quad 16 \textrm{ profiles}\\
&\Rightarrow\quad 256 \textrm{ calculations}
\end{align*}
\begin{itemize}[left=0pt]
	\item Disorder A and B require symptom 'd' and 'e' $\rightarrow$ Maximize in both
\end{itemize}
The overlapping symptoms in this case are $\{d,e\}$. Since there are no necessary symptoms for either disorder ($m^{A-} = m^{B-} = \emptyset$) it follows that $p^{*}=p^{*}_A=p^{*}_B$. Consequently, the profiles containing $p^{*}$ can be expressed as:\\

\textbf{State 1}:
\begin{align*}
\textrm{A* = [[\{d, e\}], [\{a, b, c\}, 1]]} \quad &\ldots\quad 7 \textrm{ profiles}\\
\textrm{B* = [[\{d, e\}], [\{f, g, h\}, 1]]} \quad &\ldots\quad 7 \textrm{ profiles}\\
&\Rightarrow\quad 49 \textrm{ calculations}
\end{align*}
\begin{itemize}[left=0pt]
\setlength\itemsep{-5pt}
	\item Disorder A does not have symptoms 'f', 'g' and 'h' $\rightarrow$ Minimize in B
	\item Disorder B does not have symptoms 'a', 'b' and 'c' $\rightarrow$ Minimize in A
\end{itemize}
%Minimizing these profiles is about finding the minimum cardinality from the subsets $\{a,b,c\}$ and $\{f,g,h\}$, which is directly indicated by the number $1$ in the generator, hence $a_{min} = b_{min} = 1$. As all the maximized and minimized profiles share the same size, a single instance suffices in calculating the correct similarity, so the others can be discarded. (e.g. $\{d,e,a\}$, $\{d,e,b\}$, $\{d,e,c\}$ - choose $\{d,e,a\}$) \\
Minimizing these profiles involves finding the subsets $\{a,b,c\}$ and $\{f,g,h\}$ with the smallest possible cardinality, which is indicated by the value 1 in the generator. Thus, $a_{min} = b_{min} = 1$. Since all maximized and minimized profiles share the same size, a single representative is sufficient to compute the correct similarity, and the remaining profiles can be disregarded. (e.g., among $\{d,e,a\}$, $\{d,e,b\}$, $\{d,e,c\}$ one can select $\{d,e,a\}$) \\

\textbf{State 2}:
\begin{align*}
\textrm{A** = [[\{d, e\}], [\{a\}]] = [[\{d, e, a\}]]} \quad &\ldots\quad 1 \textrm{ profile}\\
\textrm{B** = [[\{d, e\}], [\{f\}]] = [[\{d, e, f\}]]} \quad &\ldots\quad 1 \textrm{ profile}\\
&\Rightarrow\quad 1 \textrm{ calculation (result: 2/3)}
\end{align*}
In the end, only one profile (represented by a single generator (G0)) remains for each of the final conditional generators $A^{**}$ and $B^{**}$. Therefore, the MPCS$_{\max}$ computation is reduced to a single similarity calculation instead of 256.
\medskip

%\begin{align}
%M^*_A &= \{p \in M_A : 
%|p \cap p^{*}|=\max_{y \in M_A}|y \cap p^{*}| \wedge 
%|p \cap z|=\min_{y \in M^{-}_A}|p \cap y|\textrm{ for }z \in  M^{-}_A\}\\
%M^*_B &= \{p \in M_B : 
%|p \cap p^{*}|=\max_{y \in M_B}|y \cap p^{*}| \wedge 
%|p \cap z|=\min_{y \in M^{-}_B}|p \cap y|\textrm{ for }z \in  M^{-}_B\}
%\end{align}
%\begin{align}
%M^{*}_{A,p^{*}} = \argmax_{r \in M_A, s \in M^{-}_A}(|r \cap p^{*}|,-|r \cap s|)
%\end{align}
%\begin{align}
%M^*_A &= \{p \in M_A \mid \argmax_{x}|p \cap p^{\cap}| \wedge \argmin_{x \in M^{-}_A \cap %p}|x|\}\\
%P^*_B &= \{x \mid p \in M_B, \argmax_{x \in M^{\cap} \cap p}|x| \wedge \argmin_{x \in %M^{-}_B \cap p}|x|\}\\
%\end{align}
%\begin{align}
%P^*_A &= \{x \mid p \in M_A, \argmax_{x \in M^{\cap} \cap p}|x| \wedge \argmin_{x \in %M^{-}_A \cap p}|x|\}\\
%P^*_B &= \{x \mid p \in M_B, \argmax_{x \in M^{\cap} \cap p}|x| \wedge \argmin_{x \in %M^{-}_B \cap p}|x|\}\\
%\end{align}
%  &\textrm{with } X \cap p := \{m \cap p \mid m \in X\} 
\subsubsection*{Implementation}
%First, the two generators are copied to preserve the original inputs. %Formal empty sets (\{\}) in the copies are replaced with their proper definition ( set() ). 
%The symptom domain (the list of all relevant symptoms $=$ MP) is determined for both copies. The domains help to identify which symptoms can be maximized and minimized. The simplification process occurs in two stages:
%\begin{enumerate}
%\item The symptoms in the intersection of the domains that are not coming from G3 are being maximized.
%\item The symptoms not in the intersection of the domains, not coming from G3 and are missing in the other disorder are being minimized.
%\end{enumerate}
%Once both generators have undergone these optimizations, they are directly converted into the binary matrix form (AP), which serves as input for the general MPCS(max) calculation.
%\smallskip
First, both generators are duplicated to preserve the original inputs and ensure that subsequent operations do not alter the initial data. Next, the symptom domain (list of all relevant symptoms = Maximum Profile) is determined for each of the two copies. These domains serve as a basis for identifying which symptoms can be maximized or minimized during the optimization process. The simplification procedure is carried out in two distinct stages:
\begin{enumerate}
\item Maximization step: Symptoms that appear in the intersection of the two domains, and that do not originate from G3, are maximized.
\item Minimization step: Symptoms that do not belong to the intersection, are not derived from G3, and are absent in the domain of the opposite disorder, are minimized.
\end{enumerate}

\noindent After both generators have undergone these optimization steps, they are converted into their binary matrix representation (AP). This matrix serves as the input for the general MPCS$_{\max}$ computation.
\smallskip

\section{Results}
To convert the narrative descriptions in diagnostic manuals into a format that enables the automatic generation of all symptom profiles for the binary matrix representation (AP), we developed a structured approach called symptom profile generators. These generators were used to produce complete symptom profiles and to conduct initial comparisons of cognitive disorders using the MPCS algorithm.\\
%To get the narrative of diagnostic manuals into a form that allows the automatic generation of all symptom profiles for the binary matrix representation (AP), I developed this list structure called symptom profile generators. The following results are some of the generators used for producing all symptom profiles and using them to get the first results for comparing cognitive disorders with the MPCS algorithm. \\
\begin{figure}[H]
	\centering
	\fbox{\includegraphics[width=1\textwidth]{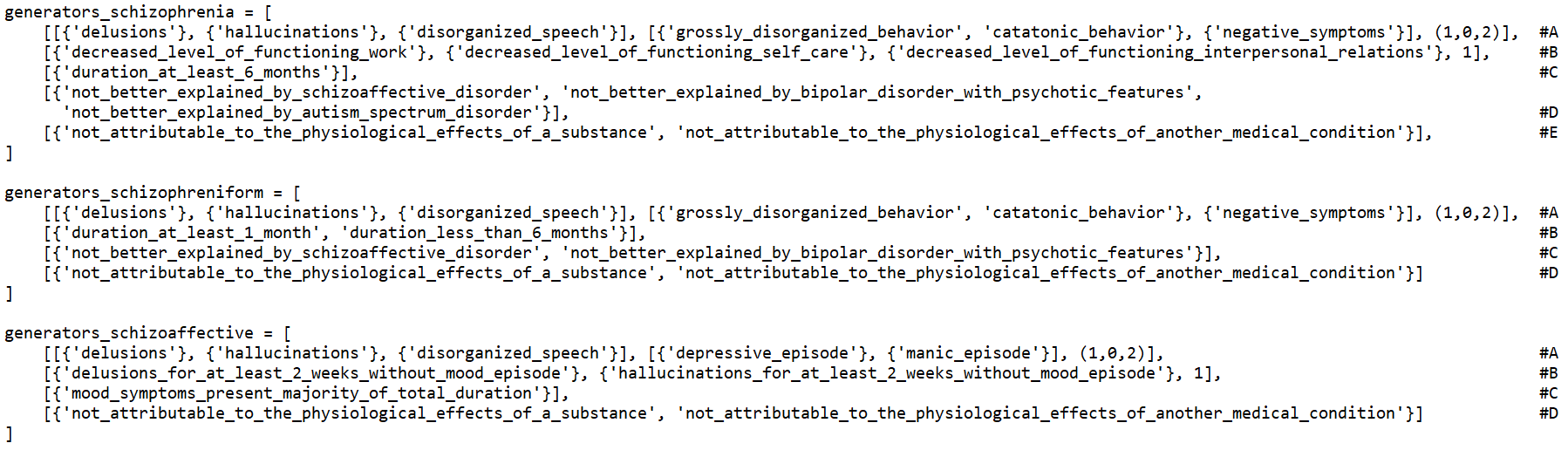}}
	\caption{Generators for Schizophrenia, Schizophreniform and Schizoaffective Disorder}
	\label{fig:6}
\end{figure}

After successfully applying the generators to simpler disorders such as schizophrenia (371 profiles), schizophreniform disorder (75 profiles), and schizoaffective disorder (53 profiles), the next step was to test them on more complex disorders. This led to the development of generators for major depressive disorder (MDD) and panic disorder, both of which required an enormous number of symptom profiles to account for all possible diagnostic pathways (MDD: 1,376,583,579 profiles, Panic disorder: 3,119,485,608 profiles).
%After the first successful results using the generators on simple disorders like "schizophrenia" (371 profiles), "schizophreniform" (75 profiles) and "schizoaffective" (53 profiles), it was time to use them for the most complex disorders the research was aiming for. This resulted in the following generators for the "major depressive disorder" (MDD) and the "panic disorder". Both of those disorders showed a huge amount of symptom profiles necessary to cover all their diagnostic pathways (MDD: 1,376,583,579 profiles, Panic: 3,119,485,608 profiles). 

\begin{figure}[H]
	\centering
	\fbox{\includegraphics[width=1\textwidth]{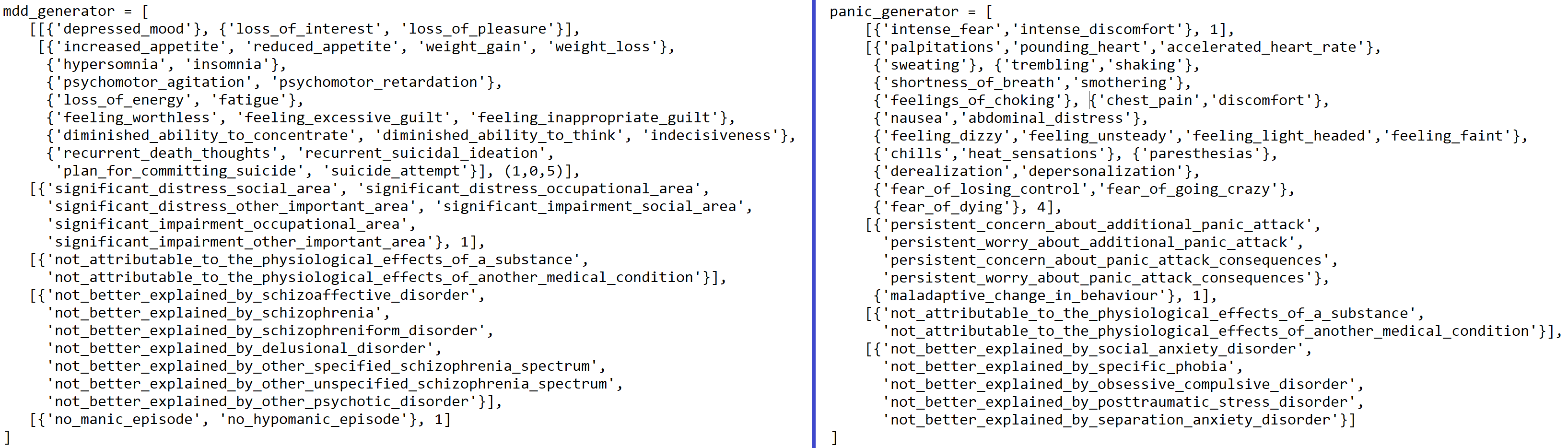}}
	\caption{Generators for Major Depressive Disorder and Panic Disorder}
	\label{fig:7}
\end{figure}

%A few other disorders were also transformed into the generator form in order to compare them in size (generator size and profile count) and for a general similarity analysis. The most useful were the "persistent depressive disorder" (PDD: 63,567 profiles) as a more manageable depressive disorder than MDD, the "generalized anxiety disorder" (27,090 profiles) as a disorder with a few of the same symptoms of other disorders and the "speech sound disorder" (7 profiles) as a totally different disorder with no relation to mood and schizophrenic tendencies.\\

Several other disorders were also converted into the generator format to compare their size (in terms of generator complexity and profile count) and to conduct a general similarity analysis. The most useful among them were:
{\footnotesize
\begin{itemize}
	\item Persistent Depressive Disorder (PDD) – 63,567 profiles,\\ providing a more manageable depressive disorder model compared to MDD.
	\item Generalized Anxiety Disorder – 27,090 profiles,\\sharing just a few symptoms with other disorders.
	\item Speech Sound Disorder – 7 profiles,\\serving as a distinct case with no relation to mood or schizophrenia-related disorders.
\end{itemize}}
%The following table are the results for the MPCS(max) calculations of a few cognitive disorders using the conditional generators, because otherwise the computations would take a very long time if even possible. The base algorithm of MPCS implemented in Python works directly with matrices up to a certain size, after that it switches to a slower vectorized version. An enhanced version of the algorithm using several techniques to improve on the base algorithm is already in the works, but still needs all the calculations done at some point, which the conditional generator approach avoids by reducing to the most similar profiles possible (only in MPCS(max)).\\
The following table presents MPCS$_{\max}$ calculations for select cognitive disorders using conditional generators, which significantly reduce computational demands. Without them, these calculations would be infeasible due to time constraints. The base MPCS algorithm, implemented in Python, operates directly on matrices up to a certain size but then switches to a slower vectorized version. While an enhanced version is in development, it still requires extensive computation. The conditional generator approach mitigates this issue by focusing only on the most similar profiles, making MPCS$_{\max}$ calculations possible for highly complex disorders.
\\

\renewcommand{\arraystretch}{1.5}
\begin{table}[H]
\begin{center}
\footnotesize
\begin{tabular}{|l|p{2.5cm}|p{2.5cm}|p{2cm}|p{2cm}|}
%\begin{tabular}{|l|c|c|c|c|}
\hline
 & \textbf{M. Depressive} & \textbf{P. Depressive} & \textbf{G. Anxiety} & \textbf{Panic}\\
\hline
\textbf{M. Depressive} & 1.000 & \textcolor{purple}{$\sim8.75 \times 10^{13}$} & \textcolor{purple}{$\sim3.73 \times 10^{13}$} & \textcolor{purple}{$\sim4.29 \times 10^{18}$}\\
\hline
\textbf{P. Depressive} & 0.763 & 1.000 & \textcolor{purple}{$\sim1.72 \times 10^{9}$} & \textcolor{purple}{$\sim1.98 \times 10^{14}$}\\
\hline
\textbf{G. Anxiety} & 0.480 & 0.519 & 1.000 & \textcolor{purple}{$\sim8.45 \times 10^{13}$}\\
\hline
\textbf{Panic} & 0.139 & 0.154 & 0.372 & 1.000\\
\hline
\end{tabular}
\vspace{3pt}
\caption{MPCS$_{\max}$ for complex disorders calculated with conditional generators. The red numbers denote the number of cosine similarity computations required in the absence of conditional generators.}
\label{table: generator_simplifier_calcs}
\end{center}
\end{table}
\vspace{-10pt}
%#mdd-pdd:    0.7627700713964737
%#mdd-panic:  0.1386750490563073
%#mdd-genanx: 0.48007935851918326
%%#pdd-panic:  0.15384615384615385
%#pdd-genanx: 0.5185449728701349
%#panic-genanx: 0.3721042037676254
%This shows that the conditional generators make it possible to calculate the MPCS of the most complex disorders with an insane amount of symptom profiles in very few calculations and basically no time, where the normal generators would have to calculate an endless quantity of cosine similarities between the profiles. \\
%The MPCS(max) of "major depressive disorder" and "panic disorder" in the table would have to compute over 4 quintillion ($4\times 10^{18}$) cosine similarities based on their number of profiles, but because of their symptoms similarities, dissimilarities and abundance of exclusive symptoms (G3), it took only 1 calculation of the two most similar profiles left in the conditional generator.
This demonstrates that conditional generators enable the calculation of MPCS for highly complex disorders with an enormous number of symptom profiles using only a few computations, making the process nearly instantaneous. In contrast, standard generators would require an impractical number of cosine similarity calculations between profiles.\\
For example, the MPCS$_{\max}$ computation for major depressive disorder (MDD) and panic disorder would typically require evaluating over 4 quintillion ($4\times 10^{18}$) cosine similarities due to their vast number of symptom profiles. Due to the conditional generator approach, which leverages symptom similarities and differences, the calculation was reduced to a single comparison between the two most similar profiles, significantly improving efficiency. This happens if no Generator 3 is involved, because the algorithm can find the two most similar profiles with a full maximization and minimization process leaving no extra profiles over (see example at the end of Sect. 2.4). In early test simulations with two matrices of size 10.000 (rows) x 50 (columns), corresponding to 100 million cosine similarities, it could take the base algorithm already several minutes (depending on the hardware) for the final result. The values from Table \ref{table: generator_simplifier_calcs} were all calculated within seconds even though their size is much larger than those earlier test simulations.
\smallskip

\section{Discussion}
%in-depth analysis\\
%interpret the results in the context of existing knowledge\\
%explain the significance of the findings\\
%address limitations of the findings\\
%suggest implications and potential future research\\

%restate the key findings in context
%emphasize the implications
%compare with existing methods

This research presents a new and necessary data representation of cognitive disorders in order to auto-generate the rows (profiles) of the machine-actionable representation with binary matrices called "All Profiles". This representation using generator lists not only makes translation from the narrative straightforward, but also yields a quick method to calculate the similarity between complex disorders that manifest in a huge amount of profiles with the MPCS. In comparison to the AP, this representation is still readable, easy to understand, can efficiently be stored and is easy to adapt/change. The conditional generators make it possible to analyse the similarity (MPCS) between all disorders in a quick fashion without the need to generate all profiles the disorders usually have, hence saving a lot of computing time. This process is limited to the MPCS$_{\max}$ method, because the value is determined by a single pair of profiles, and once this pair is located, the computation is straightforward. By contrast, MPCS(mean) depends on all profile pairs, so simplification is considerably more challenging, if feasible at all. The generator representation is not restricted to the definitions used here. These were chosen to generate profiles for the disorders from the DSM-5, but additional generators can be introduced to address other cases and research domains. Future research should explore potential applications of the generator representation and the ways generators can assist in all tasks related to the binary matrix representation of disorders. Further studies could explore improvements and optimizations of the base algorithm, following the directions in Section 2.5 (Conditional Generators).
\smallskip

\section{Conclusion}
%brief wrap-up\\
%summarize the key findings concisely\\
%reinforce the main message of the research\\
%final takeaway or recommendation\\
%------------------------------------
%summarize the main contribution
%highlight practical implications
%suggest future research directions
This research presents a new way to represent cognitive disorders via the diagnostic pathways described in their diagnostic criteria. This representation is both intuitive and adaptable, while retaining all the information required to generate the encoded enumeration of symptom profiles as a binary matrix (analogous to a disjunctive normal form), which captures the propositional logic formula described by the diagnostic criteria for further analysis. It is also expandable, has to potential to be used in other medical fields (using symptoms with diagnostic pathways) and helps with tasks regarding AP like the conditional generators to calculate the similarity (MPCS) of disorders that have far too many profiles to finish the computation in a reasonable amount of time. Future research will examine how much time and resources conditional generators save compared to standard MPCS when calculating MPCS$_{\max}$ for complex disorders, how the generators help in a recommender type system and how to generalise and improve the generator principle to make it usable in other medical fields.\\
\medskip

\section*{Data Availability}
The source data supporting this study, including the data used throughout the paper and the associated calculated values, are available in a public GitHub repository at:\\
\url{https://github.com/raoul-k/AIDA-Path/tree/main/data/generators}

\section*{Code Availability}
The code used to generate the results in this paper is available in a public GitHub repository at: \\
\url{https://github.com/raoul-k/AIDA-Path/tree/main/Python/Generators}

\section*{Declarations}
\subsection*{Abbreviations}
\noindent
\textbf{DSM}: Diagnostic and Statistical Manual of Mental Disorders \\
\textbf{ICD}: International Classification of Diseases \\
\textbf{NLP}: Natural Language Processing\\
\textbf{BM}: Binary Matrix \\
\textbf{AP}: All Profiles \\
\textbf{MP}: Maximum Profile \\
\textbf{MPCS}: Maximum Pairwise Cosine Similarity \\
\textbf{MDD}: Major Depressive Disorder \\
\textbf{PDD}: Persistent Depressive Disorder \\
\textbf{CPU}: Central Processing Unit\\
\textbf{GPU}: Graphics Processing Unit\\
\textbf{CUDA}: Compute Unified Device Architecture\\

\subsection*{Acknowledgements}
\noindent
We are grateful to Florian Hutzler (Department of Psychology, Centre for Cognitive Neuroscience, University of Salzburg) for his insightful feedback and valuable discussions throughout the development of this work. We would also like to thank our colleagues at the IDA Lab Salzburg for their support, valuable discussions, and minor contributions to this work.\\

\subsection*{Funding}
\noindent
All authors gratefully acknowledge the support of the InnovationExpress 2021 project {AIDA-PATH} (20102-F2101312-FPR). GZ gratefully acknowledges the support of the WISS 2025 projects 'IDA-Lab Salzburg' (20204-WISS/225/197-2019 and 20102-F1901166-KZP) and 'EXDIGIT' (Excellence in Digital Sciences and Interdisciplinary Technologies) (20204-WISS/263/6-6022)\\

\subsection*{Declaration of generative AI and AI-assisted technologies in the writing process}
\noindent
During the preparation of this work the author(s) used ChatGPT to improve readability and language. After using this tool, the author(s) reviewed and edited the content as needed and take(s) full responsibility for the content of the published article.\\

\appendix
\section{Full Profiles (Examples)}
\noindent
Full result (54 profiles) for G2 = [ \{a,b\}, \{c,d\}, \{e,f\}, 2 ]:
\vspace{10pt}
{\footnotesize
\begin{align*}
\textrm{2 Elements: } &\textrm{\{'a', 'e'\}}, \textrm{\{'d', 'e'\}}, \textrm{\{'b', 'e'\}}, \textrm{\{'c', 'e'\}}, \textrm{\{'a', 'd'\}}, \textrm{\{'a', 'f'\}},\\ &\textrm{\{'a', 'c'\}}, \textrm{\{'d', 'f'\}}, \textrm{\{'b', 'd'\}}, \textrm{\{'b', 'f'\}}, \textrm{\{'c', 'f'\}}, \textrm{\{'b', 'c'\}}\\
\end{align*}
\vspace{-20pt}
\begin{align*}
\textrm{3 Elements: } &\textrm{\{'a', 'd', 'e'\}}, \textrm{\{'a', 'e', 'f'\}}, \textrm{\{'a', 'b', 'e'\}}, \textrm{\{'a', 'c', 'e'\}}, \textrm{\{'d', 'e', 'f'\}},\\ &\textrm{\{'b', 'd', 'e'\}}, \textrm{\{'c', 'd', 'e'\}}, \textrm{\{'b', 'e', 'f'\}}, \textrm{\{'c', 'e', 'f'\}}, \textrm{\{'b', 'c', 'e'\}},\\ &\textrm{\{'a', 'd', 'f'\}}, \textrm{\{'a', 'b', 'd'\}}, \textrm{\{'a', 'c', 'd'\}}, \textrm{\{'a', 'b', 'f'\}}, \textrm{\{'a', 'c', 'f'\}},\\ &\textrm{\{'a', 'b', 'c'\}}, \textrm{\{'b', 'd', 'f'\}}, \textrm{\{'c', 'd', 'f'\}}, \textrm{\{'b', 'c', 'd'\}}, \textrm{\{'b', 'c', 'f'\}}\\
\end{align*}
\vspace{-20pt}
\begin{align*}
\textrm{4 Elements: } &\textrm{\{'a', 'd', 'e', 'f'\}}, \textrm{\{'a', 'b', 'd', 'e'\}}, \textrm{\{'a', 'c', 'd', 'e'\}}, \textrm{\{'a', 'b', 'e', 'f'\}},\\ & \textrm{\{'a', 'c', 'e', 'f'\}}, \textrm{\{'a', 'b', 'c', 'e'\}}, \textrm{\{'b', 'd', 'e', 'f'\}}, \textrm{\{'c', 'd', 'e', 'f'\}},\\ & \textrm{\{'b', 'c', 'd', 'e'\}}, \textrm{\{'b', 'c', 'e', 'f'\}},\textrm{\{'a', 'b', 'd', 'f'\}}, \textrm{\{'a', 'c', 'd', 'f'\}},\\ & \textrm{\{'a', 'b', 'c', 'd'\}}, \textrm{\{'a', 'b', 'c', 'f'\}}, \textrm{\{'b', 'c', 'd', 'f'\}}\\
\end{align*}
\vspace{-20pt}
\begin{align*}
\textrm{5 \& 6 Elements: } &\textrm{\{'a', 'b', 'd', 'e', 'f'\}}, \textrm{\{'a', 'c', 'd', 'e', 'f'\}}, \textrm{\{'a', 'b', 'c', 'd', 'e'\}},\\ &\textrm{\{'a', 'b', 'c', 'e', 'f'\}}, \textrm{\{'b', 'c', 'd', 'e', 'f'\}}, \textrm{\{'a', 'b', 'c', 'd', 'f'\}},\\ &\textrm{\{'a', 'b', 'c', 'd', 'e', 'f'\}}\\
\end{align*}}\\
\noindent
Full result (33 profiles) for G4 = [ [\{a,b\}, \{c\}], [\{d\}, \{e,f\}], (1,0,3) ]:
\vspace{10pt}
{\footnotesize
\begin{align*}
\textrm{3 Elements: } &\textrm{\{'c', 'd', 'e'\}}, \textrm{\{'c', 'd', 'f'\}}, \textrm{\{'b', 'd', 'e'\}}, \textrm{\{'b', 'd', 'f'\}}, \textrm{\{'a', 'd', 'e'\}},\\
& \textrm{\{'a', 'd', 'f'\}}, \textrm{\{'b', 'c', 'e'\}}, \textrm{\{'b', 'c', 'd'\}}, \textrm{\{'b', 'c', 'f'\}}, \textrm{\{'a', 'c', 'e'\}},\\
& \textrm{\{'a', 'c', 'd'\}}, \textrm{\{'a', 'c', 'f'\}}\\
\end{align*}
\vspace{-20pt}
\begin{align*}
\textrm{4 Elements: } &\textrm{\{'c', 'd', 'e', 'f'\}}, \textrm{\{'b', 'd', 'e', 'f'\}}, \textrm{\{'a', 'd', 'e', 'f'\}}, \textrm{\{'b', 'c', 'd', 'e'\}},\\
&\textrm{\{'b', 'c', 'e', 'f'\}}, \textrm{\{'b', 'c', 'd', 'f'\}}, \textrm{\{'a', 'c', 'd', 'e'\}}, \textrm{\{'a', 'c', 'e', 'f'\}},\\
&\textrm{\{'a', 'c', 'd', 'f'\}}, \textrm{\{'a', 'b', 'd', 'e'\}}, \textrm{\{'a', 'b', 'd', 'f'\}}, \textrm{\{'a', 'b', 'c', 'e'\}},\\
&\textrm{\{'a', 'b', 'c', 'd'\}}, \textrm{\{'a', 'b', 'c', 'f'\}}\\
\end{align*}
\vspace{-20pt}
\begin{align*}
\textrm{5 \& 6 Elements: } &\textrm{\{'a', 'c', 'd', 'e', 'f'\}}, \textrm{\{'a', 'b', 'd', 'e', 'f'\}}, \textrm{\{'b', 'c', 'd', 'e', 'f'\}}, \\
&\textrm{\{'a', 'b', 'c', 'd', 'e'\}}, \textrm{\{'a', 'b', 'c', 'e', 'f'\}}, \textrm{\{'a', 'b', 'c', 'd', 'f'\}}, \\
&\textrm{\{'a', 'b', 'c', 'd', 'e', 'f'\}}\\
\end{align*}}

\section{Generators}
\noindent
\textbf{Persistent Depressive Disorder}
{\scriptsize
\begin{verbatim}
[   
  [{'depressed_mood'}], #A
  [{'poor_appetite', 'overeating'}, 
   {'insomnia', 'hypersomnia'},
   {'fatigue', 'low_energy'},
   {'low_self_esteem'},
   {'poor_concentration', 'indecisiveness'},
   {'hopelessness'}, 2], #B
  [{'duration_at_least_2_years'}], #C
  Criterion #D is redundant, since it simply says "may be present"
  [{'no_past_manic_episodes', 'no_past_hypomanic_episode', 
    'no_past_cyclothymic_disorder'}], #E
  [{'not_better_explained_by_schizoaffective_disorder',
    'not_better_explained_by_schizophrenia',
    'not_better_explained_by_delusional_disorder'}], #F
  [{'not_attributable_to_the_physiological_effects_of_a_substance', 
    'not_attributable_to_the_physiological_effects_of_another_medical_condition'}], #G
  [{'significant_distress_social_area', 'significant_impairment_social_area',
    'significant_distress_occupational_area', 
    'significant_distress_other_important_area', 
    'significant_impairment_occupational_area',
    'significant_impairment_other_important_area'}, 1], #H
]
\end{verbatim}}
\noindent
\textbf{Generalized Anxiety Disorder}
{\scriptsize
\begin{verbatim}
[   
  [{'anxiety', 'worry', 'difficulty_to_control_the_worry'}], #A+B
  [{'restlessness','feeling_keyed_up','feeling_on_edge'},
   {'fatigue'},
   {'difficulty_concentrating','mind_going_blank'}, 
   {'irritability'},
   {'muscle_tension'},
   {'sleep_disturbance'}, 3], #C
  [{'significant_distress_social_area', 'significant_impairment_social_area',
   'significant_distress_occupational_area', 
   'significant_impairment_occupational_area',
   'significant_distress_other_important_area',
   'significant_impairment_other_important_area'}, 1], #D
  [{'not_attributable_to_the_physiological_effects_of_a_substance', 
    'not_attributable_to_the_physiological_effects_of_another_medical_condition'}], #E
  [{'not_better_explained_by_panic_disorder',
    'not_better_explained_by_social_anxiety_disorder',
    'not_better_explained_by_obsessive_compulsive_disorder',
    'not_better_explained_by_separation_anxiety_disorder',
    'not_better_explained_by_posttraumatic_stress_disorder',
    'not_better_explained_by_anorexia_nervosa',
    'not_better_explained_by_somatic_symptom_disorder',
    'not_better_explained_by_body_dysmorphic_disorder',
    'not_better_explained_by_illness_anxiety_disorder',
    'not_better_explained_by_schizophrenia',
    'not_better_explained_by_delusional_disorder'}]  #F
]
\end{verbatim}}
\noindent
\textbf{Speech Sound Disorder}
{\scriptsize
\begin{verbatim}
[   
  [{'persistent_difficulty_with_speech_sound_production', 
    'interference_with_speech_intelligibility', 
    'prevents_verbal_communication_of_messages'}], #A
  [{'limitations_in_effective_communication_social_participation',
    'limitations_in_effective_communication_academic_achievement',
    'limitations_in_effective_communication_occupational_performance'}, 1], #B
  [{'onset_early_developmental_period'}], #C
  [{'not_attributable_to_cerebral_palsy', 
    'not_attributable_to_cleft_palate',
    'not_attributable_to_deafness', 'not_attributable_to_hearing_loss',
    'not_attributable_to_traumatic_brain_injury',
    'not_attributable_to_other_medical_condition'}], #D
]
\end{verbatim}}
\smallskip

\bibliographystyle{unsrt}
\bibliography{generators_bibliography.bib}

\end{document}